\documentclass[runningheads]{llncs}
\usepackage[T1]{fontenc}
\usepackage{graphicx}
\usepackage[utf8]{inputenc} 
\usepackage[T1]{fontenc}    
\usepackage{hyperref}       
\usepackage{url}            
\usepackage{booktabs}       
\usepackage{amsfonts}       
\usepackage{nicefrac}       
\usepackage{microtype}      
\usepackage{xcolor}  
\usepackage{tabularx}
\usepackage{graphicx}
\usepackage{booktabs}
\usepackage{array}
\usepackage{caption}
\usepackage[table]{xcolor}

\begin{document}
\title{Eskwai for Students: Generative AI Assistant for Legal Education in Ghana}
\titlerunning{Eskwai for Students}
%

\author{George Boateng\inst{1,3} \and
Philemon Badu\inst{3}\and
Patrick Agyeman-Budu\inst{3}\and
Samuel Ansah\inst{3}\and
Evans Atompoya\inst{3}\and
Evan Igwilo\inst{3}\and
Lord Baah\inst{3}\and
Frederick Abu-Bonsrah\inst{3}\and
Victor Kumbol \inst{2,3}}


\authorrunning{G. Boateng et al.}

\institute{ETH Zurich, Switzerland\\
\and
Charité - Universitätsmedizin Berlin, Germany\\
\and
Kwame AI Inc., U.S.}


\maketitle 


\begin{abstract}
Recent advances in generative AI have shown their potential to be leveraged for legal education. Yet, work on the development and deployment of such systems for legal education in the Global South is limited. In this work, we developed Eskwai for Students, a generative AI assistant to help law students with their legal education. Eskwai for Students is a retrieval augmented generation (RAG) system that provides answers to a wide range of legal questions for law students grounded in a curated database of over 12K case laws and 1.4K legislation in Ghana. We deployed Eskwai for Students in a longitudinal study of 30 months (2.5 years) used by 3.1K law students in Ghana who made 32K queries. We evaluated the helpfulness of our AI, and provided insight into the kinds of queries law students submit to this generative AI tool, which raises some ethical concerns. This work contributes to an understanding of how law students in the Global South are using generative AI for their studies and the ways it could be leveraged responsibly to advance legal education.

\end{abstract}

\section{Introduction}
Recent advances in generative AI are reshaping legal work and education, offering new tools such as large language models (LLMs) and retrieval-augmented generation (RAG) systems that can automate cognitive tasks, summarize legal doctrines, and assist in legal research and drafting \cite{schrepel2026,surden2023}. These technologies have the potential to democratize access to legal knowledge and enhance student engagement, preparing graduates for a digitized legal profession \cite{schrepel2026,burgess2024}. However, the integration of generative AI into legal education presents unique challenges in the Global South, where disparities in technological infrastructure and access to curated legal data persist \cite{schrepel2026,hemrajani2025}.

Empirical studies in North America, Europe, and select Asian jurisdictions have begun to map the pedagogical opportunities and pitfalls of generative AI in legal education \cite{schrepel2026,regalia2024,alsbrook2024,ajevski2023,bliss2024}. These studies highlight that generative AI can effectively scaffold student understanding by providing on-demand explanations, generating hypotheticals, and serving as a platform for critical analysis and revision of legal texts. Notably, generative AI has been shown to boost student engagement, personalize learning experiences, and support the development of research and writing skills  \cite{schrepel2026}. Experiments with law students reveal that structured training in AI usage significantly enhances the quality of legal analysis and drafting, while unstructured or unguided exposure may lead to overreliance, superficial engagement, or the uncritical acceptance of AI-generated outputs \cite{schrepel2026,chen2026}.

Despite these insights, there is a significant gap in the literature addressing the specific needs and constraints of legal education in developing countries, where access to digitized legal databases is limited, and curricula are shaped by local statutes and case law, and socio-political realities that may not be reflected in the training data of mainstream generative AI models \cite{schrepel2026,hemrajani2025}. This raises critical questions about the applicability, reliability, and equity of deploying generative AI tools in these contexts.

Concerns about academic integrity, equity, and cognitive offloading are heightened in resource-constrained environments \cite{schrepel2026}. Issues such as plagiarism, unequal access to technology, and the risk of undermining rigorous legal reasoning are particularly salient \cite{schrepel2026,balan2024,farber2024}. Nevertheless, generative AI holds promise for bridging gaps in access to justice and legal knowledge, especially for marginalized students \cite{schrepel2026}.

This study addresses these gaps by evaluating "Eskwai for Students", a RAG system grounded in Ghanaian legal authorities (12K cases and 1.4K legislation) and deployed over 2.5 years with 3.1K law students. By assessing its effectiveness and analyzing student queries, this research contributes to understanding the opportunities and challenges of generative AI in legal education in the Global South. Our key contributions are (1) the design and deployment of an AI for legal education in Ghana, and (2) an evaluation of its usage and utility in a real-world learning environment in a large scale longitudinal study.

\section{Related Work}
In this section, we describe related work on using generative AI for learning support in legal education.

In a two-year randomized control experiment, undergraduate law students were assigned to three groups: prohibition, unguided exposure, and structured training to use ChatGPT and evaluated using multiple choice questions and a take-home writing assignment that required open-ended legal analysis and critical thinking. Students who had structured training had the best scores, followed by students with unguided exposure, and lastly those who were prohibited \cite{schrepel2026}. In a follow-up after a year, interestingly, the performance gap between students who used AI with and without guidance reduced  \cite{schrepel2026}. In a similarly designed study, 164 undergraduate law students were assigned to three groups: no AI assistance, optional access to AI, and 10-minute training and optional access to AI, and evaluated on an issue-spotting task. \cite{chen2026} Training significantly improved AI adoption (15\%) and examination performance (27\%).  In contrast, access without training did not improve performance. After decomposing the overall effect by principal stratification, it was found that training primarily increases the number of students who use AI rather than enhancing the scores of individual students \cite{chen2026}.  These two studies show that providing guidance or structured training improves generative AI usage and learning outcomes. However, students can develop mastery with long-term use even without initial guidance.

In contrast, some studies also report a mixed effect. For instance, in a study that evaluated 206 law students on a task that required reviewing legal complaints, it was found that AI assistance reduced the task completion time but had no significant impact on the accuracy of results \cite{nielsen2024}. Again, in another study, undergraduate law students were provided with training on the use of GPT-4 and took a second set of final exams with the aid of GPT-4. The authors found that GPT-4 assistance improved student results by 29\% in multiple-choice questions but had no effect on the essay component of the exams, which included difficult tasks like issue-spotting\cite{jonathan2023}. Interestingly, depending on the baseline results, there was a significant variation in the outcomes. The worst-performing students on the baseline test saw about 45\% improvement in scores, whereas the best-performing students saw about 20\% decline in scores \cite{jonathan2023}. These data suggest that the value of generative AI learning support may depend on the specific task and baseline performance of the students.

Lastly, Hemrajani et al. recruited 50 advanced law students to blindly evaluate the outputs of several AI models and a human expert across legal tasks, including issue spotting and summarization, drafting, legal advice, research, and reasoning in a consumer law problem \cite{hemrajani2025}. The AI models excelled in the drafting and issue-spotting tasks, either matching or surpassing the human expert. However, they underperformed in the legal research task, frequently generating hallucinations \cite{hemrajani2025}. Hallucinations are a well-known limitation of generative AI models, and it is unsurprising that the legal research task resulted in the worst performance by the AI models. RAG  is an effective strategy to mitigate this risk by providing context from relevant documents to ground the answers.

In this work, we built a RAG system grounded in a corpus of case law and legislation, an improvement over these previous works, which relied on the training data of the LLMs, a design that is less prone to hallucination. Also, unlike these previous works, we evaluated this system over an extended 2.5-year period with a large cohort of 3.1K students using the system without any limitations, providing more real-world evaluation data. By analyzing the queries made by students, we provide insights into the kinds of feedback and assistance that students expect from an AI system, which can inform the design of future systems. 

\section{System Overview}
Eskwai for Students \footnote{See the Eskwai Website: \url{https://eskwai.kwame.ai}} consists of a web app component and an AI component.

\subsection{Web App}
\begin{figure}[t]
  \centering
  \includegraphics[width=\linewidth]{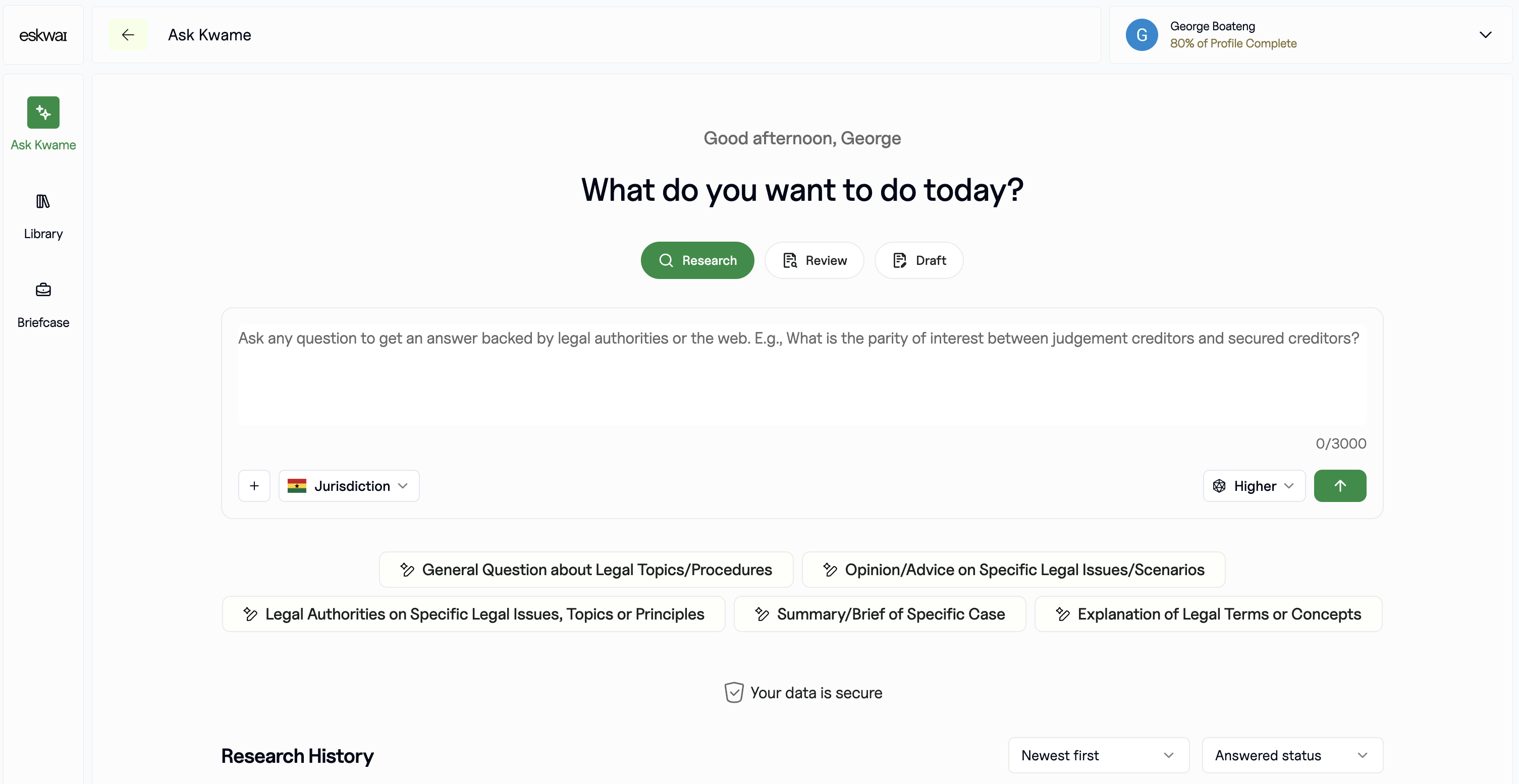}
  \caption{Screenshots of Eskwai}
  \label{fig:eskwai_webapp}
\end{figure}

\begin{figure}[h]
    \centering
    \includegraphics[width=\linewidth]{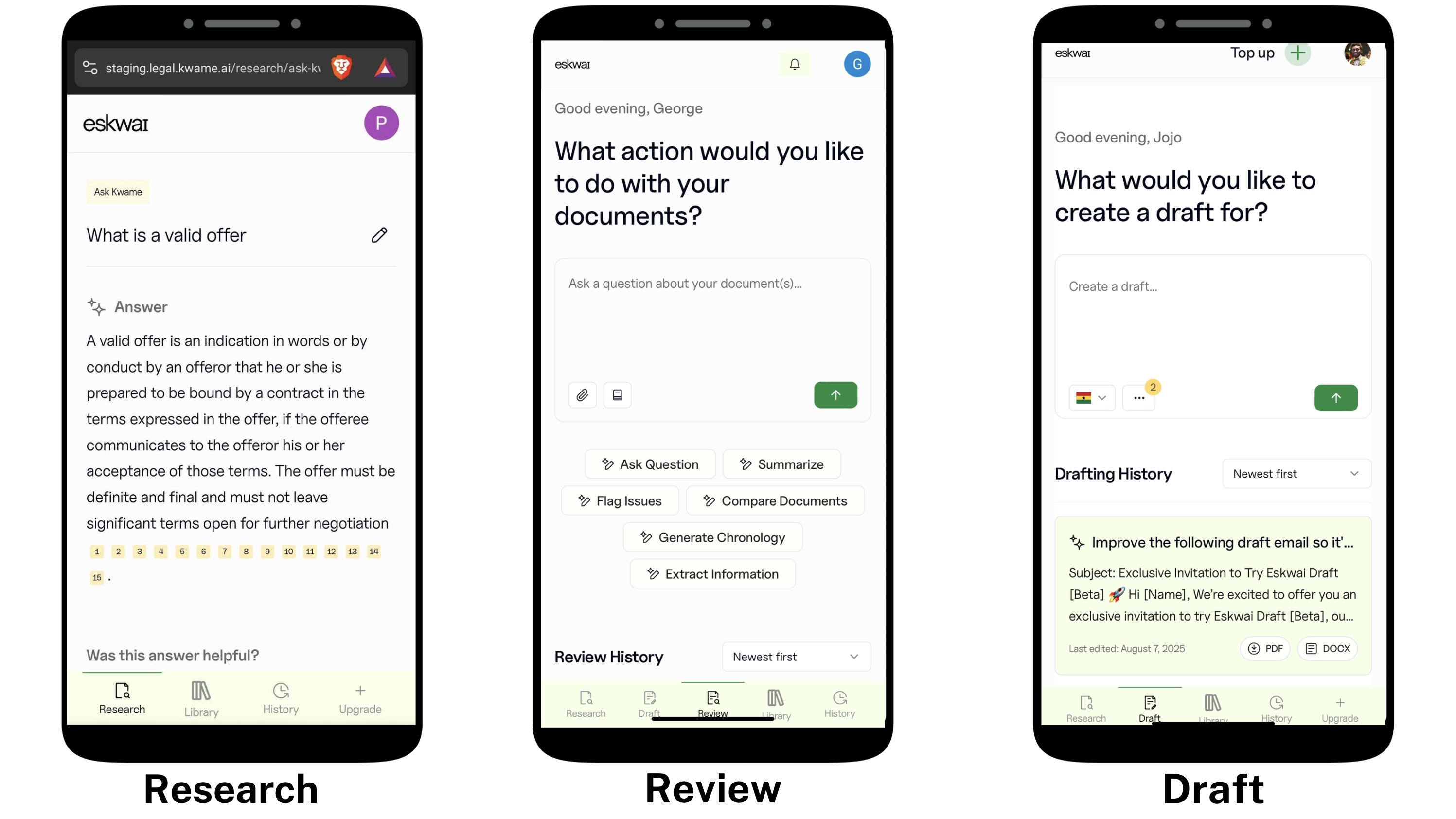}
    \caption{Screenshots of Mobile Version of Eskwai}
    \label{fig:mobile}
\end{figure}

The web app (Figure \ref{fig:eskwai_webapp}) provides a mobile-friendly interface (Figure \ref{fig:mobile}) to use Eskwai both on desktop and mobile devices. It currently consists of 3 main features: Ask Kwame, Library, and Briefcase. The Ask Kwame feature enables users to submit queries and get responses with verifiable inline citations and has 3 modes: Research, Review, and Draft. With Research, users can ask questions and get instant, trustworthy answers from case law, legislation, or the web. With review, they can analyze contracts and other legal documents by summarizing, flagging issues, and asking questions. With Draft, they can quickly generate and revise drafts of legal documents such as court processes, contracts, and correspondence.  Library enables users to view our database of cases, legislation, regulations, and templates designed for effortless navigation with intelligent search, sort, and filters. Briefcase allows users to upload, store, and organize their files in a workspace designed for effortless navigation with intelligent search, sort, and filters. These files can then be used in Ask Kwame. 

\subsection{AI Architecture}
\begin{figure}[]
  \centering
  \includegraphics[width=\linewidth]{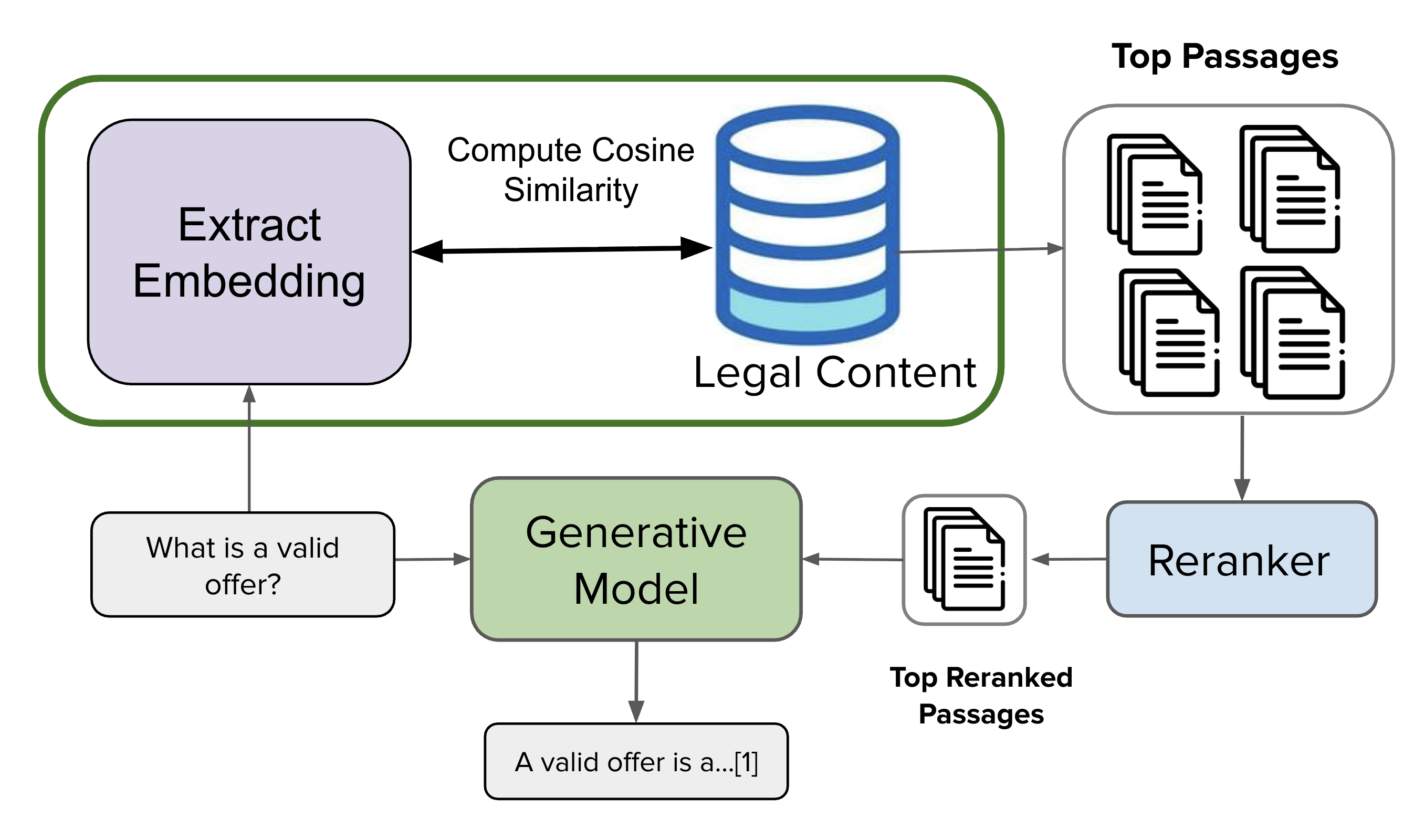}
  \caption{Overview of Eskwai RAG System}
  \label{fig:eskwai_architecture}
\end{figure}

The AI component uses a RAG system (Figure \ref{fig:eskwai_architecture})  consisting of a retrieval component that uses open-source models and a generative component that uses a commercial model (the models used have changed over the years). It generates a response to queries, grounded in a database of cases and legislation, as well as user documents, and provides inline citations to the exact passages used to generate the response. We created 5-sentence chunks from a corpus of cases (12K) and legislation (1.4K), which we curated primarily from lawyers with access to legal content. We then computed embeddings using an open-source embedding model of the chunks and saved them in ElasticSearch on a virtual machine on Google Cloud Platform.  When a query is submitted, our system first computes an embedding of the query using the same embedding model and then computes cosine similarity scores with the precomputed embeddings to retrieve the top N chunks. We then rerank the chunks using an open-source reranking model and get the top 30 chunks. We then pass the passages as context, along with a prompt, to a generative model using a GPT API from OpenAI. The answer is streamed to the user on the web app, which also shows the inline citations, which, upon clicking, open and highlight the supporting passage in the case, legislation, or user document.

\section{Deployment and Evaluation}
We launched Eskwai in September 2023, targeting lawyers and law students, publicizing mainly on social media. In this work, we evaluated usage by law students on the platform between 1st September 2023  and 28th February 2026 (30 months, 2.5 years). 

\subsection{Usability}
During the evaluation period, Eskwai recorded significant engagement from the law student community. A total of 3,127 students across more than 15 Ghanaian universities utilized the platform, collectively submitting 32,919 queries on Ask Kwame. Most users accessed Eskwai on mobile devices. Users could provide upvotes or downvotes on the answers in response to a question, "Was this helpful?", which was used to calculate a helpfulness score. User interaction was relatively limited, with 1,131 votes cast on the responses provided by the platform. Notably, only 3\% of the total queries received votes, indicating a relatively low rate of direct feedback through the voting mechanism. The platform’s helpfulness score was 68.4\%  indicating generally positive user satisfaction with room for improvement. Users who downvoted could pick from the options of possible reasons or write their own reason. Some reasons included the following: the answer did not address the specific question asked, lacked critical details such as specific cases and legislation, contained incorrect or misleading information, provided a simplistic response, provided an overcomplicated response, and contained outdated information. Analyses of the downvoted answers over the years revealed that most of them stem from (1) our database lacking specific cases and legislation referenced in queries, (2) lacking up-to-date cases and legislation, and  (3) vague and imprecise queries from users. Additionally, Eskwai demonstrated operational efficiency, with an average response latency of 7.1 seconds per query, underscoring its capacity to deliver timely legal information.

\subsection{Query Characteristics}
We analyzed the queries qualitatively to obtain broad themes and categories with the help of ChatGPT (version March 2026). Specifically, we uploaded the queries into ChatGPT and asked it to perform qualitative analyses by generating categories of queries along with examples for each. We then refined categories after reading through a random sample of queries and updated the list of representative example queries. Analysis of user queries reveals several distinct categories reflecting the academic and practical needs of law students (Table \ref{tab:query_categories}).

\textbf{Legal Opinion:} These queries involved requests for legal analysis on hypothetical or doctrinal scenarios. Some required structured responses, such as IRAC (Issue, Rule, Application, Conclusion),  or APA style and were aligned with coursework or assignment tasks. Examples include essay-length responses and critical evaluations of legal principles.

\textbf{Specific Cases and Breakdown:} These queries requested a breakdown of cases, including summaries, facts, holdings, and judicial reasoning. They sometimes only had the names of cases without additional requests.

\textbf{Procedural Explanations:} These queries focused on procedural aspects of legal practice, such as the steps involved in litigation processes.

\textbf{Definitions and Concept Clarification:} These queries involved requests for definitions of legal terminology and explanations of foundational legal concepts.

\textbf{Application of Legal Authorities:} These queries involved requests for cases or legislation to apply, support, or refute legal arguments, topics, principles or specific parts of legislation.

\textbf{Exam Question Generation:} These queries requested the generation of exam questions for studies. 

\textbf{Study Support:} These queries requested support in studies, such as drafting a study timetable or asking how to solve questions.

\textbf{Legal Drafting Tasks:} These queries involved drafting legal documents, such as court processes, agreements, or letters.

\textbf{Analysis of documents:} These queries involved performing analysis on uploaded documents, such as requesting summaries and asking questions about them.

\begin{table}
    \centering
\caption{Categories of queries along with examples}
\label{tab:query_categories}
    \begin{tabular}{|>{\raggedright\arraybackslash}p{0.3\linewidth}|>{\raggedright\arraybackslash}p{0.7\linewidth}|}\hline
         \textbf{Categories of Questions}& \textbf{Example Queries}\\\hline
         Legal opinion on scenarios, principles, or concepts& - Equity protects confidential information through a flexible set of principles, not a rigid formula”. write essay in 6 pages with detailed authorities and APA reference style \newline - With the aid of relevant authorities, demonstrate your understanding of four primary sources of law in ghana. Using IRAC
\newline - Adams Smith's cannons are archiac principles which has no place in Ghana, to what extent do you agree\\\hline
         Specific cases, including structured breakdown like briefs (facts, holding, legal principles etc.)& - AG v Sallah — Give me facts, holdings, dissenting and majority decision
\newline - Kumakye v Ghana Water and Sewage Corp
\newline - What legal principles were established in Mensah v. Adu?\\\hline
         Explanation of procedural rules& - How to prove title to land in Ghana
\newline - Can you give me a typical demonstration of how a motion is moved in court?
\newline - What are the rules of procedure in the District Court of Ghana\\\hline
         Definitions of terminologies and legal concepts& - What is injunction?
\newline - What is the meaning of partnership
\newline - difference between a licencee and an adverse possessor\\\hline
         Application of legal authorities on specific topics, legal principles or specific parts of legislation.& - I want cases to support section 19 of Companies Act, Act 992
\newline - Cases on the power of attorney whose validity date expires during or before the trial of a case
\newline - Cases that supports educational rights in Ghana\\\hline
         Generating exam questions& - Give me possible Multiple choice questions on State responsibility in Public International Law
\newline - Give some ghana legal system questions likely in exams and how you must solve it perfectly\\\hline
         Study support& - How do we use the arac in solving law questions in exams
\newline - Draft a study timetable for a law student who studies these six courses: company law, commercial law,tort law, international law, information technology law, and law and Accountable Institutions .
\newline - What do I need to know before reading law\\\hline
         Drafting legal documents & - Draft an affidavit of means for a spouse applying for maintenance
\newline - Draft a demand letter for unlawful contract termination use the Ghana labour laws
\newline - Draft a rent agreement for me\\\hline
         Analysis of documents& - Generate a summary of these document(s).
\newline - review for publication on the high street journal suggest changes
\newline - REVIEW THIS CONTRACT PER GHANA LAW POSITION\\ \hline
    \end{tabular}

\end{table}

\subsection{Discussion}
The evaluation of Eskwai’s deployment among law students highlights several key themes. The substantial number of student users and queries indicates a strong demand for AI-driven legal research tools within academic settings. The helpfulness score reflects a generally positive reception, though the low percentage of queries with votes suggests that students may be less inclined to provide explicit feedback. The platform’s rapid response time is a critical factor in its usability and overall effectiveness. Overall, the evaluation demonstrates that Eskwai supports a wide range of academic and practical legal tasks, with particular strength in analytical writing and case law assistance. However, ethical concerns remain given that students use Eskwai for generating answers to their homework problems, which could shortcut the learning process for these students. An important next step is to partner with law faculties so that we place restrictions on the type of questions our AI provides answers to, depending on the stage of the curriculum. For example, for earlier stages of their legal education, it will be important for students to perform activities like briefing cases without AI help for learning. Whereas in later stages, where they might do more application-type tasks, it might no longer be necessary to manually brief cases, but use AI for that, as they would have mastered those foundational skills.

\section{Limitations and Future Work}
In this work, we did not evaluate the impact of the tool on learning outcomes. In the future, we will perform this evaluation using various approaches such as pre/post test analysis and randomized control trial on local and national exams. 

Only a small percentage of queries received votes, which is expected in real-world settings like this. We will explore approaches to incentivize votes, such as giving AI credits for votes. 

Future work will tackle the reasons for downvotes to improve the helpfulness score, such as setting up partnerships to ensure we have comprehensive, up-to-date cases and legislation, and also train users on proper prompting approaches. One feature we recently implemented is "Magic Query," which automatically refines users' queries to be more precise and address the issue of bad prompting. We will also work on new features to enhance learning, such as test-taking and voice conversations for case simulations, and partner with law faculties to put in guardrails to address ethical concerns around the use of Eskwai by their students.

\section{Conclusion}
In this work, we developed Eskwai for Students, a generative AI assistant to help law students with their legal education. Eskwai for Students is a RAG system that provides answers to a wide range of legal questions for law students grounded in a curated database of over 12K case laws and 1.4K legislation in Ghana. We deployed Eskwai for Students in a longitudinal study of 30 months (2.5 years) in Ghana that was used by 3.1K law students who made 32K queries. Our evaluation showed that it had a decent helpfulness score of 68.4\%. Analysis of the queries showed a wide range of use cases, including doing homework problems, which raises ethical concerns around students’ usage of Eskwai. This work contributes to an understanding of how law students in Ghana are using generative AI for their studies and the ways it could be leveraged to responsibly advance legal education. 

\begin{credits}
\subsubsection{\ackname} We are grateful to the annotators who annotated the cases and legislation for Eskwai. This work was done with funding from the Afrodeutscher Startup Competition (AiDia 2023).
\end{credits}


\bibliographystyle{splncs04}
\bibliography{refs}

\end{document}